\documentclass[10pt,twocolumn,letterpaper]{article}

\usepackage{cvpr}
\usepackage{times}
\usepackage{epsfig}
\usepackage{graphicx}
\usepackage{amsmath}
\usepackage{amssymb}

\def\etal{\emph{et al}.}

\def\eg{\emph{e.g.}}
\def\ie{\emph{i.e.}}
\def\tl{\tilde{L}}
\def\argmin{\mathop{\mathrm{argmin}}}  

\usepackage{bbold}
\usepackage{bbm}
\usepackage{booktabs} 
\usepackage{array}  %
\usepackage[tight,normalsize,sf,SF]{subfigure}
\usepackage{multirow}
\usepackage{url}
\usepackage{color}
\usepackage{float}

\cvprfinalcopy 

\def\cvprPaperID{1632} 

\ifcvprfinal\fi
\begin{document}

\title{Triplet-Center Loss for Multi-View 3D Object Retrieval}

\author{Xinwei He \qquad Yang Zhou\thanks{Corresponding author.} \qquad Zhichao Zhou  \qquad Song Bai  \qquad Xiang Bai\\
Huazhong University of Science and Technology\\
\tt\small \{eriche.hust, zhouyangvcc\}@gmail.com \qquad \tt\small\{zhichaozhou,songbai,xbai\}@hust.edu.cn 
}

\maketitle
\thispagestyle{empty}

\begin{abstract}\label{sec:abstract}

Most existing 3D object recognition algorithms  focus on leveraging the strong discriminative power of deep learning models with softmax loss for the classification of 3D data,  while learning discriminative features with deep metric learning for 3D object retrieval  is more or less neglected.

In the paper,  we  study variants of deep metric learning losses for  3D object retrieval, which did not receive enough attention from this area. First , two kinds of representative losses, triplet loss and center loss,  are introduced which could  learn more discriminative features than traditional classification loss. Then, we propose a novel loss named triplet-center loss, which can further enhance the discriminative power of the features. The proposed triplet-center loss learns a center for each class and requires that the distances between samples and centers from the same class are closer than those from different classes. Extensive experimental results on two popular 3D object retrieval benchmarks and two widely-adopted sketch-based 3D shape retrieval benchmarks consistently demonstrate the effectiveness of our proposed loss, and significant improvements have been achieved compared with the state-of-the-arts. 
\end{abstract}

\section{Introduction}\label{sec:introduction}

In the past few years, 3D shape analysis has received extensive attention from both computer vision and graphics communities. Especially, many new attempts have been made to this field, thanks to the powerful deep learning approaches and the large scale 3D model benchmarks such as ShapeNet~\cite{chang2015shapenet}. 3D object retrieval is a fundamental issue in shape analysis that is the most crucial for processing and analyzing 3D data. However, most deep learning based approaches focus on leveraging the strong discriminative power of deep learning models for the classification of 3D data, \eg,~\cite{su2015multi, johns2016pairwise, qi2016volumetric}, only a few novel deep learning based approaches specifically designed for 3D object retrieval in large scale have been presented. 

With a long history in the community, 3D object retrieval may be coarsely divided into two categories: \textit{view-based} and \textit{model-based} methods. View-based methods~\cite{bai2016gift,su2015multi} extract or learn the shape features from a set of 2D view projections, where 2D convolutional neural networks (CNN) are often adopted to process such projection images. Model-based methods~\cite{furuya2016deep,wang2017cnn} obtain the 3D shape features directly from the original 3D representations so that 3D CNN is preferred. Until now, view-based methods usually outperform model-based ones in term of retrieval accuracy, as reported in the recent competitions of large scale 3D SHape REtrieval Contest (SHREC)~\cite{biasotti2014shrec, savva2016shrec, savva2017shrec}. 

One well-known example of 3D object retrieval is Multi-View Convolutional Neural Networks (MVCNN)~\cite{su2015multi}, a combination of multiple 2D projection features learned by CNN within an end-to-end trainable fashion. Similar to MVCNN, great efforts have been made to build a unified deep learning model that can simultaneously perform the tasks of 3D object classification and retrieval. These approaches including MVCNN believe that a strong classification model trained with deep learning, often can meanwhile provide a faithful similarity for 3D object retrieval.

\begin{figure*}[!htb]  \label{fig:pipeline}
\begin{center}
\includegraphics[width=0.9\linewidth]{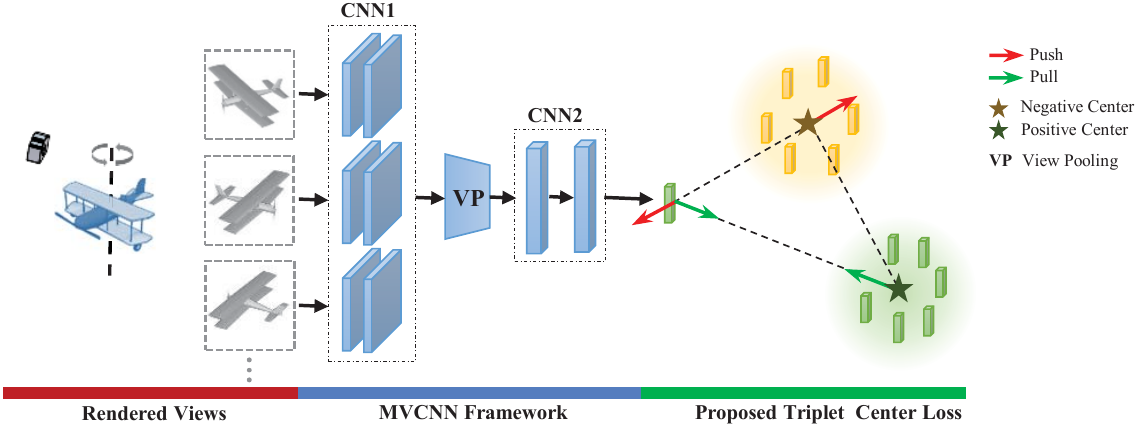}
\end{center}
\caption{
An overview of architecture for 3D  object retrieval. We adopt MVCNN as the basic component for achieving view-based 3D object representations, and the proposed TCL is used as the supervision loss. In addition, softmax loss could be also combined into the framework  for boosting performance.}
\label{fig:whole_model}
\vspace{-3ex}
\end{figure*}

In fact, deep learning approaches for 3D object retrieval are quite similar to those for image or other object retrieval, where several loss functions such as contrastive loss~\cite{chopra2005learning} 
and triplet loss~\cite{schroff2015facenet} have been introduced for training CNN, in order to learn a metric or an embedding space that makes the instances from the same category closer to each other than those from different categories. In particular, training a plain CNN model with the triplet loss for end-to-end metric learning has shown its advantages in face recognition~\cite{schroff2015facenet} and person re-identification (re-ID)~\cite{hermans2017defense}. Though the remarkable progresses in the tasks of re-ID in 2D image sets have been achieved using such loss functions, they are more or less neglected by the area of 3D object retrieval. Indeed most existing deep learning approaches for 3D shape retrieval focus on designing the sophisticated architectures of deep neural networks or exploiting different representations of 3D object. 

In contrast to most existing algorithms, in this paper we argue that training a CNN with the triplet loss~\cite{schroff2015facenet} or center loss~\cite{wen2016discriminative} that is specific for distance measure, can also bring the performance benefits to 3D object retrieval, significantly outperforming the state-of-the-art approaches on the most popular benchmark datasets of 3D object retrieval, such as ModelNet40 and ShapeNet Core55. 

In summary, we make the following contributions: 1) We firstly introduce two kinds of typical loss functions that are suggested for 3D object retrieval, and fully investigate their impact on the retrieval performance; 2) We propose a novel loss function named triplet-center loss (TCL), and show that the state-of-the-art results are obtained when using TCL based on the same CNN model, superior to other alternatives. 

The proposed TCL, motivated by center loss~\cite{wen2016discriminative} and triplet loss~\cite{schroff2015facenet}, learns a center for each class and requires that the distances between samples and centers from the same class are smaller than those from different classes, such that the features of samples from the same class are pulled closer to the corresponding center and meanwhile pushed away from the other centers of different classes. Different from center loss which only focuses on reducing the intra-class variations, 
TCL also considers inter-class separability. Compared with triplet loss, TCL avoids the complex construction of triplets and hard sample mining mechanism. 
With TCL, our CNN model for 3D object retrieval is built upon the framework of MVCNN~\cite{su2015multi}, as illustrated in Figure~\ref{fig:pipeline}.
Therefore, our method can be considered as a view-based approach that unifies the extraction of 3D shape features and distance metric learning into an end-to-end learning procedure.
Beyond ModelNet40 and ShapeNet Core55, we also demonstrate its advantages in sketch-based 3D shape retrieval task on two widely-adopted benchmarks, SHREC'13 and SHREC'14 sketch track benchmark datasets respectively.

\begin{figure*}[!tb]
\begin{center}
\includegraphics[width=\linewidth]{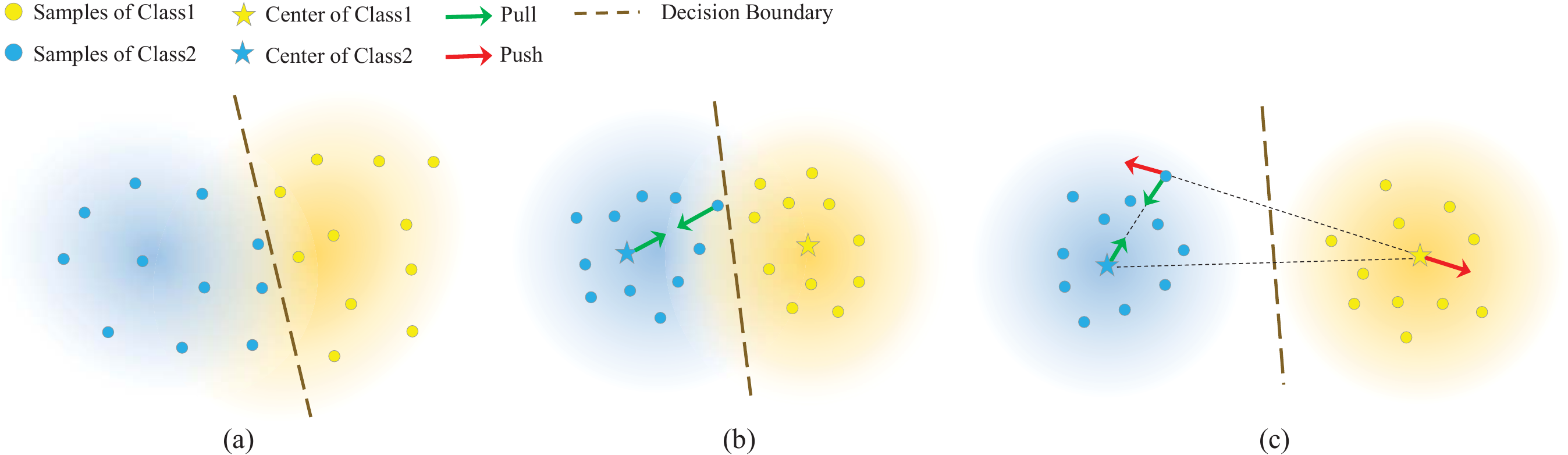}
\end{center}
\vspace{-2ex}
\caption{A toy illustration of the distributions of deep features learned by (a) softmax
loss, (b) center loss + softmax loss, and (c) triplet-center loss + softmax loss. Intuitively, the decision boundary of the softmax classifier separates the two classes elaborately. The center loss pulls features toward their corresponding centers. The TCL pulls the features to their corresponding centers and  pushes the features away from the other centers.}
\vspace{-2ex}
\label{fig:illu_example}
\end{figure*}

\section{Related work}

With the availability of large-scale labeled 3D shape collections like ShapeNet~\cite{chang2015shapenet}, a growing body of literature on 3D shape analysis especially on deep learning has emerged recently. We refer the readers to ~\cite{tangelder2008survey, ioannidou2017deep} for a comprehensive survey of 3D shape retrieval. In this section, we will mainly focus on representative 3D shape retrieval methods based on deep learning mechanisms.

In general, 3D shape retrieval methods could be roughly categorized into two classes: 3D model-based methods and view-based methods. 3D model-based methods directly learn shape features from 3D data formats, such as polygon meshes or surfaces~\cite{boscaini2016learning, BosMasRodBroCre16, xie2017deepshape, zhu2016learning}, 
voxel grid~\cite{maturana2015voxnet, wu20153d, li2016fpnn, qi2016volumetric, sedaghat2017orientation}, and point clouds~\cite{qi2017pointnet, qi2017pointnetplusplus}. For example, Furuya~\etal~\cite{furuya2016deep} propose DLAN network to process local regions of 3D shape directly and aggregate local 3D rotation-invariant features to perform retrieval task. Klokov~\etal~\cite{klokov2017escape} propose Kd-network to work with unstructured point clouds and use the learned features to perform retrieval task.  The main limitations of these methods lie in the restriction of shape representation (\eg,~smooth manifold), or high computational complexity, especially for the voxel-based methods. More recently, Wang~\etal~\cite{wang2017cnn} propose a 3D CNN based on octree representation, which can largely improve the computation efficiency compared with traditional full-voxel-based representations.

The view-based methods usually render a single view or multiple views for a 3D shape firstly, such that sophisticated image feature extractors like CNN can be exploited to extract features from the 2D rendered view, then these extracted view features are assembled  into a compact shape descriptor which is finally employed for the retrieval or classification task. For example, MVCNN~\cite{su2015multi} utilizes a max-pooling layer to aggregate the features of different views extracted by a shared CNN. Bai~\etal~\cite{bai2016gift,bai2017gift} propose a deep learning-based 3D shape search engine named GIFT, which particularly focuses on the real-time property and scalability of shape retrieval. To this end, GPU and inverted file are used to accelerate the view feature extraction and indexing, and excellent retrieval performances on various shape benchmarks are achieved. Typically, the learned features from view-based methods are more discriminative for 3D shapes, leading to a better retrieval performance in most cases. 

However, most of the above methods are not specifically designed for the 3D shape retrieval task, while in image retrieval community, in order to learn more robust and discriminative features, deep metric learning has been widely adopted. Triplet loss~\cite{schroff2015facenet}, which is proposed by Weinberger and Saul, encourages features of data points with the same identity to get closer than those with different identities. Several variants of triplet loss have also been proposed, such as \cite{liu2016deep, oh2016deep, Wang2017Angular}. However, triplet loss may suffer from the problem of time-consuming mining of hard triplets and dramatic data expansion.  Aimed at solving this problem, recently Hermans~\etal~\cite{hermans2017defense} propose a batch-hard based triplet loss (BHL) which mines hard negative and hard positive samples from the on-line training batches, and achieve state-of-the-art results on several person re-ID benchmarks. On the other hand, center loss~\cite{wen2016discriminative} has been brought up which serves as an auxiliary loss for softmax loss for the sake of learning more discriminative features for the problem of face verification. The main objective of center loss is to learn a center for the features of each class and pull features of the same class to the corresponding center more closely. 

Inspired by the success of the applications of  deep metric learning approaches for 2D image retrieval/re-ID tasks, we introduce two kinds of representative deep metric learning loss,~\ie,~triplet loss and center loss, to 3D object retrieval. In addition, a novel loss named triplet-center loss is put forward. \textcolor{black}{Recently, Wang~\etal~\cite{wang2017normface} propose a similar loss for face verification problem. However, our triplet-center loss comes from a very different intuition\footnote{\cite{wang2017normface} is motivated by normalization on weights and features, while our inspiration comes from  triplet loss and center loss.}. Besides, our loss eliminates the need for normalization on features and weights and does not reuse weights from fully connected layer.}
Remarkable improvements over the state-of-the-art  on two 3D shape retrieval benchmarks and two 3D sketch-based retrieval benchmarks demonstrate its effectiveness. 

\section{Proposed method}

In the shape retrieval task, obtaining a robust and discriminative representation of a shape is crucial for obtaining good performance. Usually, this can be partly achieved by exploiting softmax loss to train a CNN on the labeled training set. However, the learned features optimized with the supervision of softmax loss are not discriminative enough in nature, since they only focus on finding a decision boundary to separate shapes of different classes, without considering the intra-class compactness of the features. As illustrated in~Fig.~\ref{fig:illu_example} (a), although samples of the two classes are separated by the decision boundary elaborately, there exists significant intra-class variations. To cope with this problem, a lot of deep metric learning algorithms have been put forward. Here we first introduce two kinds of representative losses,~\ie,~triplet loss~\cite{schroff2015facenet} and center loss~\cite{wen2016discriminative}. Then, based on these two losses, we derive our proposed TCL.

\subsection{Review on triplet loss} \label{triplet_loss}
Triplet loss, as its name suggested, is calculated on the triplet of training samples ($x_a^i$, $x_+^i$, $x_-^i$), where  ($x_+^i$, $x_a^i$)  have the same class labels and  ($x_-^i$, $x_a^i$) have different class labels. $x_a^i$ is usually taken as an anchor of a triplet. Intuitively, triplet loss encourages to find an embedding space  where the distances between samples from the same classes (~\ie,~$x_+^i$ and $x_a^i$) are smaller than those from different classes (~\ie,~$x_-^i$ and $x_a^i$) by at least a margin $m$. Specifically, the triplet loss could be computed as follows:
\begin{equation}
\begin{split}
L_{tpl} = \sum_{i=1}^{N} \max \Big(0, m + D\big(f(x^i_a), f(x^{i}_+)\big) \\
- D\big(f(x^i_a), f(x^{i}_-)\big) \Big)
\end{split}
\end{equation}

where $f(\cdot)$ represents the feature embedding output from the neural networks, $D(\cdot)$ measures the distance between two input vectors, $N$ stands for the number of triplets in the training set and $i$ denotes the $i$-th triplet. 
However, the number of triplets grows cubically when the training dataset gets larger, which usually results in a long impractical training period. 
Moreover, the performance of triplet loss highly relies on the mining of hard triplets, which is also time consuming. Meanwhile, how to define ``good" hard triplets is still an open problem. All factors above make triplet loss hard to train. 
In order to overcome these limitations of triplet loss, we will integrate it with center loss (see Sec.~\ref{center_loss}) and propose a novel TCL loss.   

\subsection{Review on center loss}
\label{center_loss}
Center loss~\cite{wen2016discriminative} was proposed to compensate for softmax loss in face verification. It learns a center for the features of each class and meanwhile tries to pull the deep features of the same class close to the corresponding  center as illustrated in Fig.~\ref{fig:illu_example} (b). Basically, center loss can be formulated as:
\begin{equation}
L_{c} = \frac{1}{2} \sum_{i=1}^{N} D\big(f(x^i), c_{y_i}\big) 
\end{equation}
where $c_{y_i} \in \mathbb{R}^d$ is the center of class $y_i$, with $d$ denoting the dimension of features. Function $D(\cdot)$ stands for the squared Euclidean distance. During training, center loss encourages instances of the same classes to be closer to a learnable class center. However, since the parametric centers are updated at each iteration based on a mini-batch instead of the whole dataset,  which is very unstable, it has to be under the joint supervision of softmax loss during training. 

\subsection{The proposed triplet-center loss} 
\label{triplet-center-loss}

\paragraph{Motivation.} 
Though the joint supervision of center loss and softmax loss aims at minimizing the intra-class variations and achieves very promising performances on face recognition, however, as illustrated in Fig.~\ref{fig:illu_example} (b), even though the intra-class variations are very small, the inter-class clusters are very likely overlapped. This is due to that it does not consider the  separability of inter-class explicitly. While for triplet loss, it directly optimizes the network for the final task but subjects to the complexity of the construction of triplets. Motivated by the two representative losses, we bring up the triplet-center loss for the sake of learning more discriminative features efficiently.

\vspace{-2ex}\paragraph{Definitions.} 
The goal of TCL is to leverage the advantages of triplet loss and center loss,~\ie,~to efficiently minimize the intra-class distances of the deeply-learned features as well as maximize the inter-class distances of the deep features simultaneously. Let the given training dataset $\{(x^i, y^i)\}_{i=1}^{N}$ consists of $N$ samples $x^{i}\in \mathcal{X}$ with the associated labels $y^i \in \{1, 2, ...,|\mathcal{Y}|\}$. And these samples are embedded into $d$-dimensional vectors with a neural network denoted by $f_{\theta}(\cdot)$. In TCL, we assume that the features of 3D shapes from the same class share one corresponding center. Thus we can obtain $\mathcal{C} = \{c_1, c_2, ..., c_{|Y|}\}$, where $c_{y} \in \mathbb{R}^d$ denotes the center vector for samples with label $y$, and $|\mathcal{Y}|$ is the number of centers. For simplicity, we adopt $f_i$ to represent $f(x^i)$ in the following paper. Similar to center loss, we update the parametric centers at each iteration based on a mini-batch. Given a batch of training data with $M$ samples, we define TCL as 
\begin{equation}
\label{eq:tcl}
L_{tc}=\sum_{i=1}^{M} \max\Big(D\big(f_i, c_{y^i}\big) + m - 
\min_{j\neq y^i} D\big(f_i,c_{j}\big), 0\Big)
\end{equation}
where $D(\cdot)$ represents the squared Euclidean distance function denoted as:
\begin{equation}
D(f_i, c_{y^i}) = \frac{1}{2}||f_i - c_{y^i}||_2^2 \\
\end{equation}
As illustrated in Fig.~\ref{fig:illu_example} (c), TCL is to push the distances between the samples and their corresponding center $c_{y^i}$ closer than the distances between the samples and their nearest \emph{negative center} (~\ie,~centers of the other classes $\mathcal{C} \setminus \{c_{y^i}\}$) by a margin $m$.

To compute the back-propagation gradients of the input feature embeddings and the corresponding centers, we assume the following notations for demonstration: $\mathbb{1}{[\emph{condition}]}$ is an indicator function which outputs $1$ if the \emph{condition} is satisfied and outputs $0$ otherwise, $q_i =~\argmin_{j\neq~y^i}D\big(f_i, c_{j}\big)$ is an integer index which indicates the nearest negative center of $i$-th sample, and $\tl_i$ represents the triple-center loss of $i$-th sample as
\begin{align}
\tl_i=\max\Big(D\big(f_i, c_{y^i}\big) + m - \min_{j\neq y^i} D\big(f_i,c_{j}\big), 0\Big)
\end{align}
Then, the derivatives of our TCL loss Eq.~\ref{eq:tcl} with respect to the feature embedding of $i$-th sample $\frac{\partial L_{tc}}{\partial f_i}$ and $j$-th center $\frac{\partial L_{tc}}{\partial c_j}$ can be calculated as follows:
\begin{equation}
\begin{split}
\frac{\partial L_{tc}}{\partial f_i} &= \Big(\frac{\partial D\big(f_i, c_{y^i}\big)}{\partial f_i} - \frac{\partial D\big(f_i, c_{q_i}\big)}{\partial f_i}\Big)\cdot ~ \mathbb{1}{[\tl_i > 0]} \\
&= (c_{q_i}-c_{y^i})\cdot~\mathbb{1}{[\tl_i > 0]}
\end{split}
\end{equation}

\begin{equation}
\begin{split}
\frac{\partial L_{tc}}{\partial c_j} = \frac{\sum_{i=1}^{M}
(f_i - c_j)\cdot \mathbb{1}{[\tl_i > 0]}\cdot \mathbb{1}{[y^i = j]} }{1+\sum_{i=1}^{M}\mathbb{1}{[\tl_i >0]} \cdot \mathbb{1}{[y^i = j]}} \\
- \frac{\sum_{i=1}^{M}(f_i-c_j)\cdot \mathbb{1}{[\tl_i > 0]} \cdot \mathbb{1}{[q_i = j]}}{1+\sum_{i=1}^{M}\mathbb{1}{[\tl_i > 0]} \cdot \mathbb{1}{[q_i = j]}}
\end{split}
\end{equation}

\paragraph{Joint supervision with softmax loss.} 
Softmax loss focuses on mapping the samples to discrete labels,  while TCL aims to apply metric learning to the learned embeddings directly. Unlike center loss, TCL can be used independently from softmax loss. However, these two losses can also be combined together to achieve more discriminative and robust embeddings according to our experiments in Sec.~\ref{sec:exp}, which can be written as 
\begin{equation}
\label{eq:ensemble}
    L_{total} = \lambda L_{tc} +  L_{softmax}
\end{equation}
where $\lambda$ is a hyper-parameter which controls the trade-off between the TCL and softmax loss. We attribute the benefit brought by softmax loss to the fact that the parametric centers of TCL are randomly initialized and updated based-on the mini-batches instead of the whole datasets which might be tricky, while softmax loss could serve as a good guider for seeking better class centers.
\subsection{Discussion} \label{Sec:Disc}

\paragraph{Compared with triplet loss.}
Different from triplet loss, whose triplet is made up of triple samples $(x_a^i, x_+^i, x_-^i)$, the triplet of the TCL is composed of the $i$-th sample $x^{i}$, its corresponding center $c_{y^i}$ and its nearest \emph{negative center}. For a training dataset with $N$ samples, only $N$ triplets will be formed for TCL while the number of triplets for triplet loss is $O(N^3)$ much more than TCL. Consequently, in comparison with  triplet loss, TCL avoids the complexity of constructing triplets and the necessity for mining hard samples.
We provide an empirical analysis on the two losses and visualize the learned embeddings using t-SNE in Sec.~\ref{visualization}. 


\vspace{-2ex}\paragraph{Compared with center loss.} TCL can be taken as 
a variant of triplet loss, which could be exploited as supervision for training the neural networks independently of softmax loss, while center loss has to be combined with softmax to make the learning feasible otherwise the deeply learned features and centers will degrade to zeros according to~\cite{wen2016discriminative}. Furthermore, TCL simultaneously maximizes the intra-class compactness and inter-class separability explicitly while center loss neglects the latter one which may lead to inter-class overlapping. Besides, center loss aims at reducing the absolute distances between the samples and their corresponding centers, while TCL penalizes the relative distances with a hinge-style loss which is more relaxed and easier to train. For empirical analysis of TCL loss and center loss, see Sec.~\ref{visualization} for details.

\section{Experiments}\label{sec:exp}
In this section, we evaluate the performance of the proposed TCL on two kinds of 3D shape retrieval tasks: \textit{generic 3D shape retrieval task} and \textit{sketch-based 3D shape retrieval task}. The former is a within-domain retrieval task, where the given query and examples in the database are 3D models. While the latter is a cross-domain retrieval task, the given query are 2D sketches instead. 

\subsection{Generic 3D shape retrieval task}
\paragraph{Datasets.} We evaluate our TCL loss on two well-known 3D shape benchmarks: ModelNet40~\cite{wu20153d} and ShapeNet Core 55~\cite{savva2016shrec}. The ModelNet40 dataset contains 12,311 CAD models from 40 common categories. For this dataset, we follow previous works~\cite{su2015multi,bai2016gift} on the training and testing split settings,~\ie,~randomly selecting 100 models per category, of which 80 models are used as training data and the rest for testing. The ShapeNet Core 55 dataset~\cite{savva2016shrec} is composed of 51,190 3D shapes in total from 55 categories, which are further divided into 204 sub-categories. Due to its diversity of categories and large variations within classes, ShapeNet55 dataset is quite challenging. The whole dataset is split into the training set, the validation set and the test set, which contains 35,765, 5,159, 10,266 models, respectively. Further, this dataset has two variants (ShapeNet 55 perturbed dataset and ShapeNet 55 normal dataset). For the normal dataset, shapes are aligned. For perturbed version of the dataset, each model is randomly rotated by an angle. 
We conduct experiments on the perturbed dataset which is more challenging. All retrieval performances are reported on the testing set.

\vspace{-2ex}\paragraph{Implementation Details.} \label{para:implement}
Experimental codes are implemented in PyTorch~\url{http://pytorch.org/} and executed on a server with four Nvidia Titan-X GPUs, an Intel i7 CPU and 64GB RAM. We choose the Imagenet-pretrained VGG-A with batch normalization~\cite{simonyan2014very} \footnote{\small Different from the original MVCNN, which adopts VGG-M as their base network, we adopt VGG-A with batch normalization to validate our proposed loss due to the lack of pre-trained VGG-M in Pytorch.} as our base network. VGG-A contains 8 convolution layers (\textit{conv 1-8}) with kernels of size 3 by 3 and 3 fully-connected layers (\textit{fc 9-11}). And the view-pooling layer is placed right after \textit{conv 6}. The layers before and after the view-pooling layer are denoted as CNN1 and CNN2 respectively. We initialize the centers with a Gaussian distribution, and the mean and standard deviation is (0, 0.01) respectively. For optimization, we adopt the stochastic gradient descent algorithm with a mini-batch size of 16 for all our experiments. The learning rate for optimizing CNN1 and CNN2 is set to be 1e-4 and 1e-3 respectively. And for both CNN1 and CNN2, a weight decay of 1e-4 and a momentum of 0.9 are used. \textcolor{black}{The learning rate for centers is set to 0.1 for our experiments}. We clip the gradient of the center by 0.01. During testing, We extract the output of \textit{fc-10}, which is 4096-dimensional, as the features for all our retrieval tasks.

\vspace{-2ex}\paragraph{Parameter influence.} As indicated by loss function in~Eq.~\eqref{eq:ensemble}, the margin parameter $m$ and  $\lambda$ may affect the final combinations of the losses. Specifically, $\lambda$ in~Eq.~\eqref{eq:ensemble} controls the trade-off between softmax loss and TCL loss, while $m$ controls the relative distance between the sample embeddings to its corresponding center and to its nearest negative center. To study the impact of the two hyper-parameters, we give an empirical analysis on ModelNet40 dataset. 

\begin{figure}[!ht]
\begin{center}
\includegraphics[width=\linewidth]{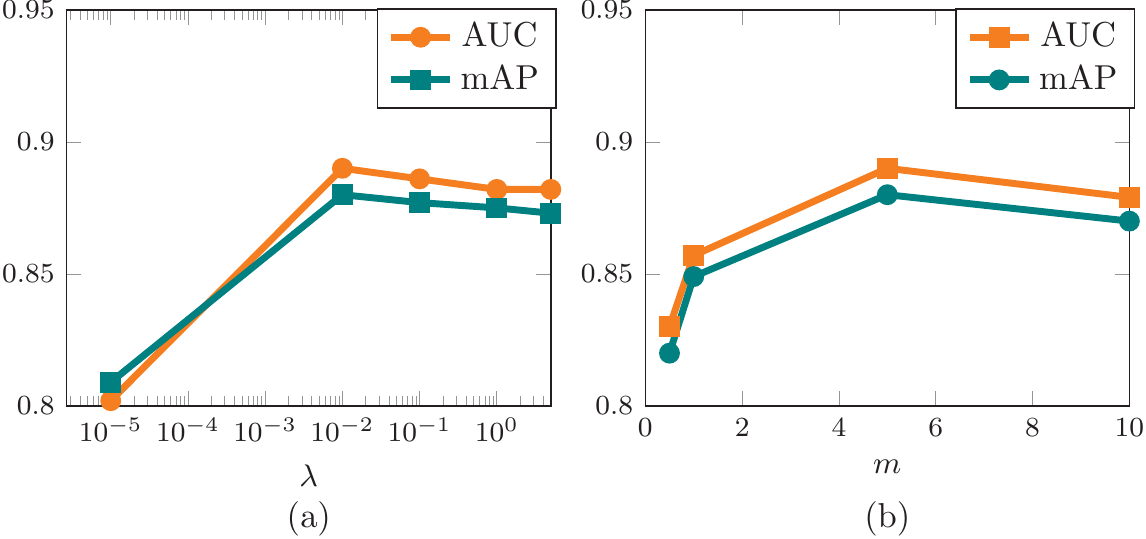}
\end{center}
\vspace{-2ex}
\caption{The retrieval performances achieved by (a) varying $\lambda$ when $m$ is fixed to 5 and (b) varying $m$ when $\lambda$ is fixed to 0.01.}
\label{fig:param_influence}
\vspace{-2ex}
\end{figure}

The influence of hyper-parameter $\lambda$ is presented in Fig.~\ref{fig:param_influence} (a). 
From the experimental results,  we can see that TCL is very robust to this parameter.  For a wide-ranged values from 0.01 to 10, the trained models consistently achieves very promising results on ModelNet40 dataset. We assume that this is because TCL and softmax loss are  complementary  losses. TCL focuses on the feature  representation directly, while softmax loss focuses on how to separate or map the feature representation into a discrete label space. Once the model with softmax loss has converged,  TCL can further enforce the feature embeddings into more compact clusters without sacrificing the classification performance too much, thus good retrieval performance can be gained, and vise versa. When $\lambda$ is set to be 0, which means the model is trained using only softmax loss, the performance is worst, only achieving an mAP of nearly 80.0\%. But with TCL, we can get an improvement of 7$\sim$8\% in terms of mAP. However, TCL loss is more sensitive to the margin $m$. To study the influence of $m$, we fixed $\lambda$ to be 0.01, and then set $m$ to be 0.5, 1, 5 and 10 respectively. For hyper-parameter $m$, too large or too small values both lead to inferior results. When it is too small, the strength of triplet center loss will be weakened, while too large value may cause over-fitting problem. The best results are achieved by setting $m$ to be 5. This gives an mAP of  88.0\% and an AUC of 89.0\%. Thus, we set $m$ and $\lambda$ to be 5 and 0.01 in default respectively for the following experiments. 

\vspace{-2ex}\paragraph{Comparison with other losses.}  \label{differentloss}
To validate the proposed TCL loss, we conduct extensive experiments on various losses, including triplet loss, softmax loss, softmax loss with center loss, TCL and softmax loss with TCL on ModelNet40 dataset. As can be seen from Tab.~\ref{table_1}, TCL and softmax loss with TCL perform best among these losses, obtaining an mAP of 86.7\% and 88.0\% respectively. In addition, softmax loss with center loss can increase the mAP by 3.4\% over softmax loss, reaching  83.5\%. The results of triplet loss are merely comparable to softmax loss. 

\begin{figure*}[!ht]
\begin{center}
\includegraphics[width=\linewidth]{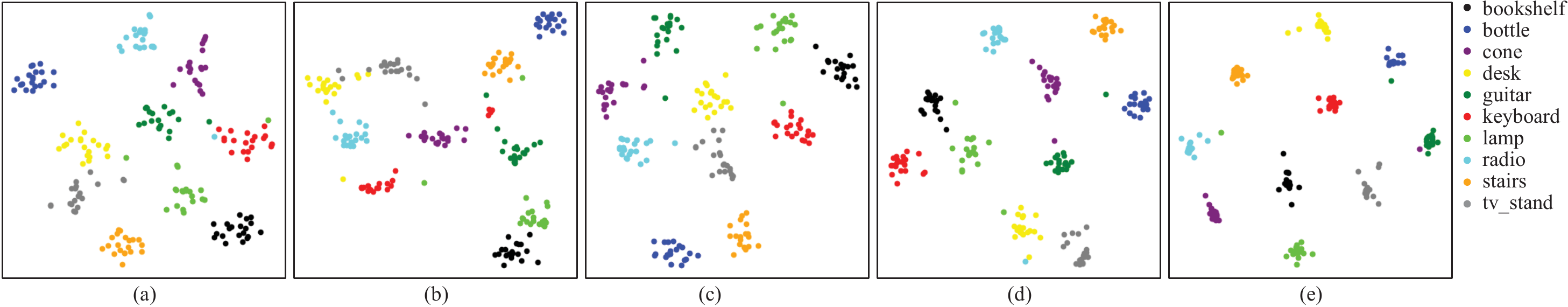}
\end{center}
\caption{A visualization of deeply-learned features by (a) softmax loss, (b) triplet loss, (c) softmax loss + center loss, (d) triplet center loss, (e) softmax loss + triplet center loss. Here we randomly select 10 classes from the test set for visualization (Best viewed in color).}
\label{fig:vis}
\vspace{-2ex}
\end{figure*}

\vspace{-2ex}\paragraph{Visualization of learned representations.} \label{visualization}
We adopt t-SNE~\cite{maaten2008visualizing} to visualize the deeply-learned features of the samples from ModelNet40 dataset. As is shown in Fig.~\ref{fig:vis}, some nice properties can be observed: (i) Compared with softmax loss, the learned embeddings of the same class are obviously getting closer to each other after introducing center loss, while triplet loss also results in slightly better embeddings since the sample embedding distances are considered directly.  (ii) The proposed TCL performs better than center loss on achieving small intra-class variance and large inter-class variance. (iii) With the combination of TCL + softmax loss, learned embeddings make the most compact and separated clusters at the same time. This demonstrates the intuition that better underlying representations for retrieval task can be obtained with the proposed TCL loss.

\begin{table}[tb]
\small
\caption{The performances (\%) of different losses on ModelNet40.}
\vspace{1ex}
\label{table_1}
\centering
\begin{tabular}{lp{1.9cm}<{\centering}p{1.9cm}<{\centering}p{1.9cm}<{\centering}}
\toprule Loss function & AUC & MAP  \\
\midrule
        softmax loss & 80.9 & 80.2  \\ 
        softmax+center loss & 84.3 & 83.5 \\
        triplet loss    & 80.9 & 80.0 \\
	    TCL & 87.6 & 86.7  \\
        TCL+softmax loss & \textbf{89.0} & \textbf{88.0} \\
\bottomrule
\end{tabular}
\vspace{-2ex}
\end{table}

\begin{figure}[!ht]
\begin{center}
\includegraphics[width= \linewidth, height=1.8in] 
{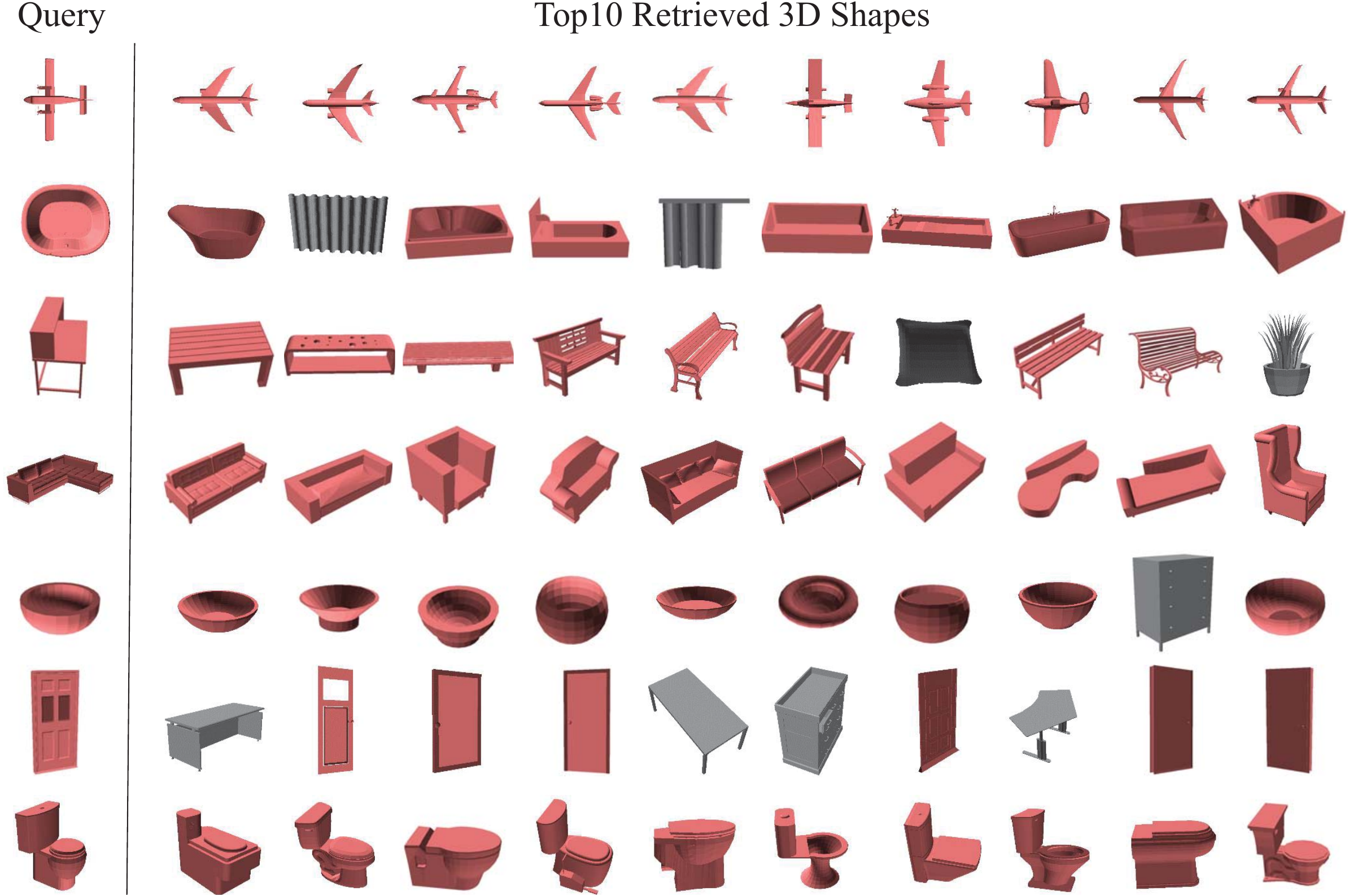}
\end{center}
\caption{Retrieval examples on ModelNet40 dataset. Top matches are shown for each query with mistakes highlighted in gray.}
\label{fig:shape_results}
\end{figure}

\vspace{-2ex}\paragraph{Comparison with the state-of-the-arts.} The performance comparisons with the state-of-the-art shape retrieval methods are presented in Tab.~\ref{comp_modelnet40_other_methods}. Here we choose view-based retrieval methods including MVCNN~\cite{su2015multi}, GIFT~\cite{bai2016gift}, DeepPano~\cite{shi2015deeppano}, and voxel-based retrieval methods including 3DShapeNet~\cite{notchenko2016sparse}, DLAN~\cite{furuya2016deep} for comparison. As is shown, compared with MVCNN, which is firstly trained with softmax loss and then an off-line large-margin metric learning algorithm is applied, our method (TCL+softmax loss) has demonstrated its superior discriminative ability and improved mAP by nearly 7\%, reaching 88.0\%. Furthermore, we outperform GIFT by nearly 6\% in terms of AUC and mAP respectively. And compared with DLAN, which is the current state-of-the art method on ModelNet40, an improvement of 3\% in terms of mAP is obtained. Besides, our method also beats a recent re-ranking algorithm named Regularized Ensemble Diffusion (RED)~\cite{Bai_2017_ICCV}. Fig.~\ref{fig:shape_results} shows the representative retrieval results of our method on the ModelNet40 dataset.  

\begin{table}[tb]
\small
\caption{The performance (\%) comparison with state-of-the-art methods on ModelNet40. '-' represents missing metric.}
\vspace{2ex}
\label{comp_modelnet40_other_methods}
\centering
\begin{tabular}{lp{1.9cm}<{\centering}p{1.9cm}<{\centering}p{1.9cm}<{\centering}}
\toprule Methods & AUC & MAP  \\
\midrule
        3DShapeNet~\cite{notchenko2016sparse}  & 49.9 & 49.2 \\
        DeepPano~\cite{shi2015deeppano} &  77.6 & 76.8 \\ 
        MVCNN~\cite{su2015multi} &  - & 80.2  \\ 
         GIFT~\cite{bai2016gift} & 83.1 & 81.9 \\
        DLAN~\cite{furuya2016deep} & - & 85.0  \\ 
        RED~\cite{Bai_2017_ICCV} & 87.0  & 86.3 \\
	  Ours & \textbf{89.0} & \textbf{88.0} \\
\bottomrule
\end{tabular}
\vspace{-2ex}
\end{table}
\vspace{-2ex}\paragraph{Experiments on ShapeNet Core55.} We evaluate our proposed TCL on the more challenging ShapeNet Core55 dataset. Since each shape in the dataset is attached with a category label and a subcategory label. To accurately evaluate the more fine-grained retrieval results, the organizer of SHREC'16 uses a metric named normalized discounted cumulative gain (NDCG). 
In addition, other traditional evaluation metrics like F-measure and mAP are also used. 
The comprehensive comparisons with the state-of-the-art methods on the perturbed dataset are listed in Tab.~\ref{shrec_table2}, which is extremely challenging for the existing methods. We choose the following methods for comparisons, including GIFT~\cite{bai2016gift}, MVCNN~\cite{su2015multi}, Wang~\cite{savva2016shrec}, Li~\cite{savva2016shrec}, Kd-network~\cite{klokov2017escape}.
As is shown, our method (TCL+softmax loss) outperforms the strong baseline MVCNN by nearly 10\% in terms of the micro-averaged mAP and 12\% for the macro-averaged mAP. For other metrics, we also outperform by a large margin. Further, compared with GIFT, the former state-of-the-art in the contest, notable improvements are also achieved. 
These results demonstrate that the proposed loss is good at obtaining robust and discriminative representations for 3D shapes once again. 
\begin{table}[tb]
\small
 \caption{The performance (\%) comparison on ShapeNet55 test perturbed.}
 \label{shrec_table2}
 \vspace{1ex}
\centering
\begin{tabular}{lp{0.56cm}<{\centering}p{0.56cm}<{\centering}p{0.60cm}<{\centering}p{0.56cm}<{\centering}p{0.56cm}<{\centering}p{0.56cm}<{\centering}p{0.56cm}}
\toprule
\multirow{2}{*}{Methods} & \multicolumn{3}{c}{microALL}   & \multicolumn{3}{c}{macroALL}  \\
\cmidrule(lr){2-4} \cmidrule(l){5-7}
    & F1 & mAP & NDCG  & F1 & mAP & NDCG \\
        \midrule
         MVCNN~\cite{su2015multi} & 61.2 & 73.4 & 84.3  & 41.6 & 66.2 & 79.3  \\
        GIFT~\cite{bai2016gift}  & 66.1 & 81.1 & 88.9  & 42.3 & 73.0 & 84.3   \\
        Wang~\cite{savva2016shrec} & 24.6 &  60.0 & 77.6  & 16.3 & 47.8 & 69.5 \\ 
        
        Li~\cite{savva2016shrec}   &  53.4 & 74.9 & 86.5  & 18.2 & 57.9 & 76.7 \\ 
        Kd-network~\cite{klokov2017escape}  & 45.1 & 61.7 & 81.4 & 24.1 & 48.4 & 72.6   \\
        Ours  & \textbf{67.9} & \textbf{84.0} & \textbf{89.5} & \textbf{43.9} & \textbf{78.3} & \textbf{86.9} \\
\bottomrule
\end{tabular}
\vspace{-3ex}
\end{table}

\subsection{Sketch-based 3D shape retrieval task}
Considering the huge visual gap between sketch and 3D shape, sketch-based shape retrieval is a quite challenging problem. Here we further demonstrate the potential advantages of TCL in such a cross-domain retrieval task.
\vspace{-2ex}\paragraph{Datasets.} Two widely adopted sketch-based shape retrieval benchmarks have been adopted to evaluate our proposed method, they are SHREC'13 and SHREC'14 sketch track benchmark datasets respectively. SHREC'13 is composed of 7,200 sketches and 1,258 shapes, which are grouped into 90 classes. There are 80 sketches for each class, of which 50 sketches for training and 30 for testing. SHREC'14, which is the largest sketch-based shape retrieval dataset currently, contains 13,680 sketches and 8,987 shapes, divided into 171 classes. Due to its large variations within category and diversity of classes, SHREC'14 dataset is very challenging. For each class of sketches, 50 sketches are selected for training and 30 for testing.

\vspace{-2ex}\paragraph{Implementation Details.} As for the experimental setup, we follow the same training-testing split as ~\cite{Li2013SHREC,xie2017learning}. Similar to ~\cite{xie2017learning}, we render 12 views for each shape and employ two AlexNets~\cite{alex2012imageNet} to extract features of 2D projections and sketches, respectively. MVCNN framework is adopted to obtain the representations of 3D shapes. In addition, we assume that the features from these two different domains share the same centers and classifiers. Thus we could train the two AlexNets pre-trained on ImageNet dataset jointly via back propagation with TCL and softmax loss. The parameters are given as ${\lambda}$ = 0.01, $m$ = 5.0. The other settings are the same as Sec.~\ref{para:implement}. To measure the retrieval performances, we use the following metrics including Nearest Neighbor (NN), First Tier (FT), Second Tier (ST), E-measure (E), Discounted Cumulated Gain (DCG) and mean Average Precision (mAP).

\vspace{-2ex}\paragraph{Experiments on SHREC'13 and SHREC'14.} In Tab.~\ref{table_shrec13} and Tab.~\ref{table_shrec14}, a comprehensive comparison with various state-the-of-arts is presented, including 
Sketch-Based Retrieval method with View Clustering (SBR-VC)~\cite{Li2013SHREC}, Cross Domain Manifold Ranking method (CDMR)~\cite{furuya2013ranking}, Siamese network (Siamese)~\cite{wang2015sketch}, Deep Correlated Metric Learning (DCML)~\cite{dai2017deep} and Learned Wasserstein Barycentric Representations (LWBR)~\cite{xie2017learning}. Among them, DCML and LWBR perform best. 
DCML~\cite{dai2017deep} proposes discriminative loss and correlation loss, aiming to increase the discrimination of features within each domain as well as the correlation between different domains. While LWBR~\cite{xie2017learning} mainly focuses on improving 3D shape representations which proposes to adopt wasserstein barycenters of features of multiple views as 3D shape representations instead of max-pooling in MVCNN~\cite{su2015multi}. These methods have advanced other methods a lot thanks to the powerful of deep learning. Especially LWBR gets 75.2\% mAP on SHREC'13, nearly 8\% higher than DCML. However, benefit from TCL, more discriminative features are obtained. Even without sophisticated computation of barycentric representations, our method (TCL+softmax loss) achieves 79.8\% mAP on SHREC'13 dataset, 46.2\% mAP on SHREC'14 dataset, which outperforms LWBR with at least 5\% on both datasets. 
\begin{table}[tb]
	\small
    \centering
    \renewcommand{\arraystretch}{1}
    \caption{The performance (\%) comparison on SHREC’13 dataset}
    \vspace{1ex}
    \label{table_shrec13}
    \begin{tabular}{lp{0.58cm}<{\centering}p{0.58cm}<{\centering}p{0.58cm}<{\centering}p{0.58cm}<{\centering}p{0.58cm}<{\centering}p{0.58cm}<{\centering}p{0.58cm}<{\centering}}
    \toprule
        Methods  & NN & FT & ST & E & DCG & mAP \\
	\midrule
        CDMR~\cite{furuya2013ranking}  & 27.9 & 20.3 & 29.6 & 16.6 & 45.8 & 25.0 \\
        SBR-VC~\cite{Li2013SHREC} & 16.4 & 9.7 & 14.9 & 8.5 & 34.8 & 11.6 \\
        Siamese~\cite{wang2015sketch} & 40.5 & 40.3 & 54.8 & 28.7 & 60.7 & 46.9 \\
        DCML~\cite{dai2017deep} & 65.0 & 63.4 & 71.9 & 34.8 & 76.6 & 67.4 \\
        LWBR~\cite{xie2017learning} & 71.2 & 72.5 & 78.5 & 36.9 & 81.4 & 75.2 \\
        Ours & \textbf{76.3} & \textbf{78.7} & \textbf{84.9} & \textbf{39.2} & \textbf{85.4} & \textbf{80.7} \\
        \bottomrule
    \end{tabular}
\end{table}

\begin{table}[tb]
	\small
    \centering
    \renewcommand{\arraystretch}{1}
    \caption{The performance (\%) comparison on SHREC’14 dataset}
    \vspace{1ex}
    \label{table_shrec14}
    \begin{tabular}{lp{0.58cm}<{\centering}p{0.58cm}<{\centering}p{0.58cm}<{\centering}p{0.58cm}<{\centering}p{0.58cm}<{\centering}p{0.58cm}<{\centering}p{0.58cm}<{\centering}}
    \toprule
        Methods  & NN & FT & ST & E & DCG & mAP \\
	\midrule
        CDMR~\cite{furuya2013ranking}  & 10.9 & 5.7 & 8.9 & 4.1 & 32.8 & 5.4 \\
        SBR-VC~\cite{Li2013SHREC} & 9.5 & 5.0 & 8.1 & 3.7 & 31.9 & 5.0 \\
        Siamese~\cite{wang2015sketch} & 23.9 & 21.2 & 31.6 & 14.0 & 49.6 & 22.8 \\
        DCML~\cite{dai2017deep} & 27.2 & 27.5 & 34.5 & 17.1 & 49.8 & 28.6 \\
        LWBR~\cite{xie2017learning} & 40.3 & 37.8 & 45.5 & 
23.6 & 58.1 & 40.1 \\
        Ours & \textbf{58.5} & \textbf{45.5} & \textbf{53.9} & \textbf{27.5} & \textbf{66.6} & \textbf{47.7} \\
        \bottomrule
    \end{tabular}
    \vspace{-2ex}
\end{table}


    

\subsection{View-based v.s. Model-based}
So far we have demonstrated the superiority of the proposed TCL for view-based and sketch-based shape retrieval tasks. Actually, a sketch can be considered as a "special" view of the object, which is why we train them jointly with the projected views and got notable rewards in the retrieval. Whereas it has been proved that deeply-learned features from model-based approaches are complementary with those of view-based methods, we still conduct a comparison by further combining TCL into two representative model-based approaches: PointNet~\cite{qi2017pointnet} and VoxNet~\cite{maturana2015voxnet}. Experiments are performed on ModelNet40 dataset. As we can see in Tab.~\ref{table_pv},  though less impressive compared to view-based mechanisms,  significant improvements of 3\%$\sim$4\% are gained compared with baseline methods. Nevertheless, we believe that better performance can be achieved once the baseline of model-based deep learning approaches is improved. We leave it as a future work.
\begin{table}[tb]
\small
\caption{The performance (\%) comparison on ModelNet40 using VoxNet and PointNet.}
\vspace{1ex}
\label{table_pv}
\centering
\begin{tabular}{lp{0.7cm}<{\centering}p{0.7cm}<{\centering}p{0.7cm}<{\centering}p{0.7cm}<{\centering}}
\toprule
\multirow{2}{*}{Methods} & \multicolumn{2}{c}{VoxNet}   & \multicolumn{2}{c}{PointNet}  \\
\cmidrule(lr){2-3} \cmidrule(l){4-5}
    & AUC & MAP & AUC & MAP \\
\midrule
softmax loss & 70.9 & 70.0 & 71.4 & 70.5 \\
softmax loss + center loss  & 73.1 & 72.0 & 74.3 & 73.1 \\
TCL loss & 73.4 & 72.4 & 72.3 & 71.3 \\
TCL + softmax loss & \textbf{74.3} & \textbf{73.2} & \textbf{75.6} & \textbf{74.5} \\
\bottomrule
\end{tabular}
\vspace{-2ex}
\end{table}

\section{Conclusion} \label{conclusion}
In this paper, we focus on the view-based 3D shape retrieval task and propose a novel loss function named triplet center loss. This loss function combines the advantages of triplet loss and center loss, and has demonstrated its effectiveness on both 3D shape retrieval task and sketch based shape retrieval task. The triplet center loss can optimize the features directly by minimizing the intra-class variance while also maximizing the inter-class variance at the same time. As a result, the learned embeddings are more robust and discriminative, thus more appropriate for retrieval task. 
In the future, we would like to explore our proposed loss to other retrieval tasks such as person re-identification, face recognition and content-based image retrieval.  
\paragraph{Acknowledgements} This work was supported by NSFC 61573160 and 61602461, to Dr. Xiang Bai by the National Program for Support of Top-notch Young Professionals and the Program for HUST Academic Frontier Youth Team.

{\small
\bibliographystyle{ieee}
\bibliography{references}
}

\end{document}